\documentclass[11pt,a4paper]{article}
\usepackage[hyperref]{acl2021}
\usepackage{times}
\usepackage{latexsym}
\usepackage{amsmath}
\usepackage{graphicx}
\usepackage{makecell}

\usepackage{microtype}

\aclfinalcopy 

\title{Ontology-Enhanced Slot Filling}

\author{Yuhao Ding and Yik-Cheung Tam\\
  Department of Computer Science \\
  NYU Shanghai \\
  1555 Century Avenue, Pudong New District, Shanghai, China 200122 \\
  \texttt{\{yd1158,yt2267\}@nyu.edu}}

\date{}

\begin{document}
\maketitle
\begin{abstract}

Slot filling is a fundamental task in dialog state tracking in task-oriented dialog systems.
In multi-domain task-oriented dialog system, user utterances and system responses may mention multiple
named entities and attributes values. A system needs to select those that are confirmed by the user and fill them into destined slots.
One difficulty is that since a dialogue session contains multiple system-user turns, feeding in all the tokens into a deep model such as BERT can be challenging due to limited capacity of input word tokens and GPU memory.
In this paper, we investigate an ontology-enhanced approach by matching the named entities occurred in all dialogue turns using ontology. 
The matched entities in the previous dialogue turns will be accumulated and encoded as additional inputs to a BERT-based dialogue state tracker. In addition,
our improvement includes ontology constraint checking and the correction of slot name tokenization. Experimental results showed that our ontology-enhanced 
dialogue state tracker improves the joint goal accuracy (slot F1) from 52.63\% (91.64\%) to 53.91\% (92\%) on MultiWOZ 2.1 corpus.

\end{abstract}

\section{Introduction}
Dialog state tracking (DST) aims at tracking users' intention and status of slot-value pairs in multi-turn dialogues. In reality, these dialogues are multi-domain and task-oriented. For example, a user may book a hotel immediately after purchasing a train ticket, completing two tasks in hotel and train domains. 
Multi-domain scenario can make DST more difficult due to the complexity of an user utterance with multiple intents and slots. Another challenge is that DST needs to comprehend relevant previous dialogue turns when updating the current dialogue state. For hotel booking, a dialogue system may recommend multiple hotel options subject to user's requirements. Then a user may decide which hotel to select or none of them can fulfill the needs.

One popular corpus for studying multi-domain task-oriented DST is MultiWOZ 2.0~\cite{budzianowski2018multiwoz} covering 7 domains ranging from hotel, attractions to train and taxi. Recently, it has been upgraded to MultiWOZ 2.1~\cite{eric2019multiwoz} after fixing annotation errors. MultiWOZ not only provides labeled dialogue states for each turn in a dialogue session, but also comes with ontology where entities such as restaurant names are augmented with attributes such as area, price range etc. We conjecture that ontology would be useful for DST because it helps identifying novel entities that are not occurred in training dialogues. Moreover, attributes of an entity provide useful information to validate consistency of filled slot-value pairs. For instance, a dialogue state tracker tries to fill in a restaurant name that violates the area constraint provided by a user. Ideally, DST should resolve the conflicts accordingly.



In this paper, we start with the recent state-of-the-art SOM-DST (Selectively Overwriting Memory for Dialogue State Tracking) that generates slot values from utterances in a dialogue context and selectively overwrites previous dialog states~\cite{kim2019efficient}. SOM-DST is based on BERT pre-trained model~\cite{devlin-etal-2019-bert}, encoding dialog utterances from one previous turn and the current turn, along with the dialog states to be filled. Our contributions in this paper is to enhance DST using ontology. First, we detect and accumulate named entities that occurred in previous dialogue turns to bring over long-distance information to the current turn that may have been missed by modeling mistakes for better natural language understanding. Second, we leverage the structured entity-attribute information to validate and correct slot-filling errors as a post-processing step. Lastly, we correct tokenization errors of some uncommon slot names which yields improved slot filling accuracy. Altogether, our approaches improve the joint goal accuracy (and slot F1) from 52.63\% (91.64\%) to 53.91\% (92.00\%) on MultiWOZ 2.1 corpus.

\section{Related work}

Earlier approaches for DST rely on a pre-defined candidate list and formulate slot filling as a classification task~\cite{henderson2014word, mrkvsic2016neural, liu2017end, zhong2018global} where each candidate in the list is considered as a class label. In practice, it is difficult to enumerate all possibilities especially on open classes like restaurant names~\cite{xu2018end}. Thus, more recent approaches extract slot values directly from dialogue context using the copying mechanism~\cite{gao2019dialog, chao2019bert, lee2019sumbt, wu2019transferable, zhang2019find, kumar2020ma}. With the effectiveness of BERT~\cite{devlin-etal-2019-bert}, these DST models employ BERT to encode dialogue context. There are also approaches to handle out-of-vocabulary and rare words using a triplet copy mechanism~\cite{heck2020trippy}.

Our approach is based on SOM-DST~\cite{kim2019efficient} that uses BERT to jointly encode dialog utterances in two turns (the previous turn and the current turn) and the previous dialogue states. The input format starts with [CLS] token followed by two dialog turns' utterances separated with [SEP]. Then each slot-value pairs from the previous dialogue states are appended as inputs.
The BERT encoder produces hidden states at each input token position. First, [CLS] hidden vector is used for domain classification. Second, hidden vector at each slot name position is used to predict an operation, namely 'CARRYOVER', 'UPDATE', 'DELETE' and 'DONTCARE'. If the predicted slot operation is 'CARRYOVER', then the corresponding value of the slot will be brought to the next turn. If the predicted operation is 'UPDATE', the model will update the slot value using pointer generator \cite{gu2016incorporating, see2017get} mechanism, either by copying the slot values directly from the dialogue turn or generating them using Gated Recurrent Unit~\cite{bahdanau2014neural}.

Prior work to improve SOM-DST includes~\citet{zhu2020efficient} that considers relations among different domain slots and constructs a schema graph with prior knowledge. The resulting graph was fed into BERT and achieved 53.19\% joint goal accuracy in MultiWOZ 2.1 corpus. Our work differs from the previous approach that we use ontology to extract and accumulate entities from previous dialogue turns and also employ a rule-based post-correction step to validate inconsistent slot-value pairs. Our approaches are easy to implement and has shown solid improvement in joint goal accuracy.

\section{Proposed approach}

We observe that the SOM-DST baseline~\cite{kim2019efficient} has rooms for improvement with the following aspects:
\begin{itemize}
    \item Lack of ontology information: The model does not utilize ontology which contains structured domain information such as restaurants, hotels, attractions. These information can be injected into the model and for post-validation.
    \item Counter-intuitive tokenization: In SOM-DST, slot names are fed as regular input tokens. However, the default BERT tokenizer on some slot names would produce counter-intuitive results. For instance, a slot name 'pricerange' is tokenized as \text{'price', '\#\#rang', '\#\#e'}. Another common slot value 'dontcare' is tokenized as \text{'don', '\#\#tc', '\#\#are'}. From a human viewpoint, these results are counter-intuitive and BERT may find these tokens difficult to understand in terms of attending them with the dialogue utterances.
\end{itemize}

Considering the above problems, we propose the following approaches for improvement.

\begin{figure*}[h!]
\centering
  \includegraphics[scale=0.5]{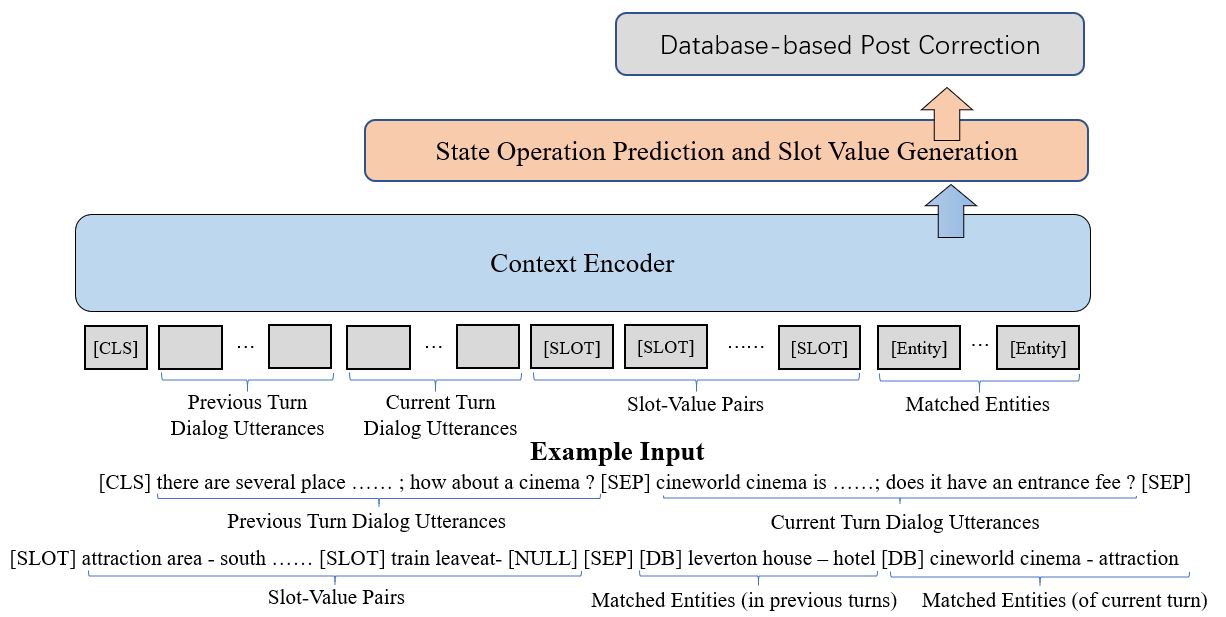}
  \caption{Proposed DST system architecture leveraging ontology in various stages.}
\label{fig:somdst}
\end{figure*}

\paragraph{Ontology-based entity enhancement}
Since ontology contains named entities and attributes such as price range, area etc, we apply simple string matching to extract named entities from previous dialogue turns that are beyond the current dialogue turn. The extracted named entities provide an informative and yet compressed way to represent the long-distant dialogue histories. Together with the extracted entities from the current dialogue turn, we append the extracted entities as additional inputs, resulting in an ontology-enhanced SOM-DST as illustrated in Figure~\ref{fig:somdst}. Sophisticated named entity recognizer can also be applied here as our future work.
Specifically, we introduce an additional '[DB]' token to format the additional named entity inputs. For instance, a restaurant 'prezzo' occurs in previous dialogue turn. Then we append '[DB] prezzo - restaurant' as additional word inputs after all the slot-value pairs. These additional DB entries do not use position embeddings. We add the extracted entities from all previous dialogue turns in an accumulative manner, bringing all unique entities into the current turn for DST.

\paragraph{Ontology-based post-correction} From error analysis, some slot-filling errors can be avoided via validation of named entity and its attributes from ontology. For example, 'the gardenia' restaurant has an 'expensive' price range in ontology. If the DST model tracks 'the gardenia' restaurant at the current turn but also predicts a 'moderate' price range, then conflict happens. Therefore, our goal is to avoid this error via ontology-based post-correction.

\paragraph{Correcting counter-intuitive tokenization} Tokenization of slot names 'pricerange' and 'dontcare' produces counter-intuitive segmented results. Therefore, we add them as additional vocabularies in the BERT tokenizer to avoid this.



\section{Experiments}
We used MultiWOZ 2.1 corpus~\cite{eric2019multiwoz} for experiments. We trained our SOM-DST baseline~\cite{kim2019efficient} using the code available from the GitHub repository~\footnote{https://github.com/clovaai/som-dst}. We implemented our proposed approaches on top of the SOM-DST baseline and reported their combination effects on slot accuracy (SA), slot F1, and joint goal accuracy (JGA) on the test set.
We mostly followed the default hyper-parameters from the code base for model training~\footnote{hyper-parameters: batch size = 16, encoder learning rate = 4e-5, decoder learning rate = 1e-4, warmup proportion (encoder and decoder) = 0.1, dropout rate = 0.1, optimizer = BertAdam}. We also followed the original SOM-DST paper and only used five domains (restaurant, train, hotel, taxi, attraction), not using hospital and police. Therefore, the input slots covered 5 domains and 30 slots for each turn. We also used the same pre-processing script by ~\citet{wu2019transferable}, which was also used by SOM-DST.

\subsection{Results}

Table~\ref{tbl:overal_results} shows the evaluation results on the MultiWOZ 2.1 test set after applying the proposed approaches. First, we obtained results for SOM-DST baseline using the out-of-the-box training settings with 256 input tokens. Next, we increased the maximum sequence length to 384 tokens and observed improvements on joint goal accuracy, slot accuracy and slot F1. After applying tokenization fix on slot names 'pricerange' and 'dontcare', we further improved SA and slot F1 but a slight hurt on JGA. Since JGA is a strict metric that requires all slots in a dialogue session to be filled correctly, it is possible that there is a slight drop in JGA even though SA and slot F1 were improved consistently. After applying ontology-based enhancement by accumulating named entities from previous user utterances, we observed an absolute 1.09\%, 0.07\%, and 0.34\% improvement on JGA, SA and slot F1 respectively compared to the baseline. These were also translated into 4\% relative reduction in slot error rate compared to the baseline. Lastly, ontology-based post-correction yielded 53.91\%, 97.38\% and 92\% on JGA, SA and slot F1 respectively. This equates to absolute 1.28\%, 0.08\% and 0.36\% improvement on JGA, SA and slot F1 respectively compared to the baseline.

Below are the best configuration to improve our DST model: 
\begin{itemize}
    \item Correct tokenization on 'pricerange' as a new word in BERT tokenizer.
    \item Apply ontology-based named entity extraction and accumulation from previous dialogue turns.
    \item Apply ontology-based post-correction on conflicting entities and their attributes.
\end{itemize}

\begin{table}
\centering
\begin{tabular}{lccc}
\hline \textbf{Name} & \textbf{JGA} & \textbf{SA}  & \textbf{Slot F1}  \\ \hline
Baseline (256) & 52.10\% & 97.26\% & 91.40\% \\
Baseline (384) & 52.63\% & 97.30\% & 91.64\% \\
\makecell[l]{+ Tokenization\\ ~~~~Fix} & 52.48\%& 97.33\% & 91.80\% \\
\makecell[l]{+ Ontology Entity\\ ~~~~Enhancement} & 53.72\%& 97.37\% & 91.98\% \\
\makecell[l]{+ Ontology\\ ~~~~Post-Correction} & \textbf{53.91\%} & \textbf{97.38\%} & \textbf{92.00\%} \\
\hline
\end{tabular}
\caption{Test set performance on MultiWOZ 2.1 after applying the proposed approaches. JGA: Joint goal accuracy. SA: Slot accuracy.}
\label{tbl:overal_results}
\end{table}

\subsubsection{Post-correction error analysis}
\begin{figure}[h!]
  \centering
  \includegraphics[scale = 0.4]{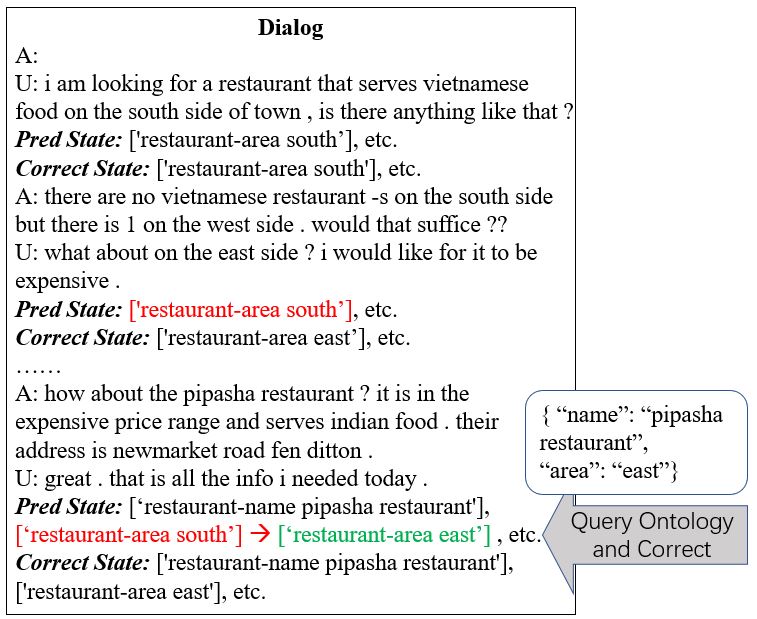}
  \caption{Example of post-correction after querying the ontology database.}
  \label{fig:error_fix}
\end{figure}

Figure~\ref{fig:error_fix} shows how the post-correction works in a dialogue sample. For example, there is a prediction error on the restaurant area due to the change of user intention but our model cannot track the change correctly. At the last turn, our model tries to fill in a restaurant name which is unambiguous from the dialogue context. Therefore, we use the ontology information of 'pipasha restaurant at the east area' and correct the wrongly predicted area attribute accordingly.

Although this is a rule-based approach, we find that it is effective on restaurant domain. In particular, the method fixes 16 'restaurant-pricerange' slots, 15 'restaurant-area' slots and 10 'restaurant-food' slots. However, we find that it is not as effective on hotel domain. Though it could correct 3 'hotel-area' slots and 5 'hotel-internet' slots, it introduces 14 errors for 'hotel-stars' slots. We believe that our model has a lower prediction accuracy for 'hotel-name', and a wrongly predicted name will likely bring an erroneous correction, deserving a better validation strategy in our future work.



\section{Conclusions and future work}
We have presented ontology-based approaches to enhance slot filling performance. Our results have shown that ontology is useful to improve dialogue state tracking. First, the accumulated named entities from previous dialogue turns are beneficial. Second, a simple rule-based post-correction can fix errors with the help of entity-attribute information from ontology. Finally, intuitive tokenization of some slot names matters to boosting slot filling accuracy.


In the future, we will explore using ontology on train domain to detect and correct inconsistent departure and arrival times and destinations. Moreover, we would like to explore incorporation of contextualized representation of named entities from previous dialogue turns.
Another future work is to identify typos and variations of restaurant and attraction names for better entity linking to ontology database. Lastly, we will evaluate the proposed ontology enhancement to other DST models.

\bibliographystyle{acl_natbib}



\end{document}